# Modelling Data Dispersion Degree in Automatic Robust Estimation for Multivariate Gaussian Mixture Models with an Application to Noisy Speech Processing

Dalei Wu[*1], Haiqing Wu[2]

Department of Electrical and Computer Engineering, Concordia University
1455 Maisonneuve Blvd. West, Montreal, QC H3G 1M8, Canada

[*1]daleiwu@gmail.com; [2]haiqingwu@gmail.com

*Abstract*

The trimming scheme with a prefixed cutoff portion is known as a method of improving the robustness of statistical models such as multivariate Gaussian mixture models (MG-MMs) in small scale tests by alleviating the impacts of outliers. However, when this method is applied to real-world data, such as noisy speech processing, it is hard to know the optimal cut-off portion to remove the outliers and sometimes removes useful data samples as well. In this paper, we propose a new method based on measuring the dispersion degree (DD) of the training data to avoid this problem, so as to realise automatic robust estimation for MGMMs. The DD model is studied by using two different measures. For each one, we theoretically prove that the DD of the data samples in a context of MGMMs approximately obeys a specific (chi or chi-square) distribution. The proposed method is evaluated on a real-world application with a moderately-sized speaker recognition task. Experiments show that the proposed method can significantly improve the robustness of the conventional training method of GMMs for speaker recognition.

*Keywords*

*Gaussian Mixture Models; K-means; Trimmed K-means; Speech Processing*

## Introduction

Statistical models, such as Gaussian Mixture Models (GMMs) [17] and Hidden Markov Models (HMMs) [15], are important techniques in many signal processing domains, which include, for instance, acoustical noise reduction [20], image recognition [8], speech/speaker recognition [15, 19, 25], etc. In this paper, we study a robust modelling issue regarding GMMs. This issue is important, since GMMs are often used as fundamental components to build some more complicated models, such as HMMs. Thus, the method studied in this paper will be useful for other models as well.

The standard training method for GMMs is Maximum likelihood Estimation (MLE) [2] based on the Expectation Maximisation (EM) algorithm [4]. Though it has been proved effective, this method still lacks robustness in its training process. For instance, it is not robust against gross outliers and cannot compensate the impacts from the out-liers contained in a training corpus. As it is well known, outliers often widely exist in a training population, due to the clean data often either being contaminated by noise, or interfered by the objects other than the claimed data. Out-liers in the training population may distract the parameters of the trained models to inappropriate locations [3] and can therefore break the models down and result in poor perform-ance of a recognition system. In order to solve this problem, a partial trimming scheme is introduced by Cuesta et al [3] to improve the robustness of the statistical models, such as K-means and GMMs. In their method, a prefixed proportion of data samples is removed from the training corpus and the rest of the data are used for model training. This method has been found robust against gross outliers when it is applied to small scale data examples.

However, when this method is used in real-world appli-cations, such as noisy acoustic signal processing, it is hard to know the optimal cutoff proportion that should be used, since one does not know what faction of the data should be taken away from the overall training population. A bigger or smaller removal proportion would result in either removing too much useful information or having no effect on remov-ing outliers. To attack this issue, we propose a new method by using a dispersion degree model with two different distance metrics to identify the outliers automatically.

The contributions of our proposed method are three-fold: First, we suggest to use the trimmed K-means algorithm to replace the conventional K-means approach to initialise the parameters of GMMs. It is showed in this paper that appropriate initial values for model parameters are crucial for the robust training of GMMs. Second, we propose to use the dispersion degree of the training data samples as a selection criterion for automatic outlier removal. We theoretically prove that the dispersion degree approximately obeys a certain distribution, depending on the measure it uses. We refer this method as the automatic robust estimation with a trimming scheme (ARE-TRIM) for Gaussian mixture models hereafter.

Third, we evaluate the proposed method on a real-world application with a moderately-sized speaker recognition task. The experimental results show that the proposed method can significantly improve the robustness of the conventional training algorithm for GMMs by making it more robust against gross outliers.

The rest of this paper is organised as follows: In Section 2, we present the framework of our proposed ARE-TRIM training algorithm for GMMs. In Section 3, we present the trimmed K-means clustering algorithm and compare it with the conventional K-means. In Section 4, we propose the dispersion degree model based on two distance metrics and use it for ARM-TRIM. We carry out the experiments to evaluate ARE-TRIM in Section 5 and finally we conclude this paper with our findings in Section 6.

## Framework of Automatic Trimming Algorithm for Gaussian Mixture Models

The proposed ARE-TRIM algorithm essentially includes several modifications to the conventional training method of GMMs. The conventional GMM training algorithm norm-ally consists of two steps: the K-means clustering algorithm for model parameter initialisation[1] and the EM algorithm for parameter fine training, as illustrated in Fig. 1. It is well known that appropriate initial values for model parameters crucially affect the final performance of GMMs [2]. Inapp-ropriate initial values could cause the fine training with EM at the second step to be trapped at a local optimum, which is often not globally optimal. Therefore, it is interesting to study more robust clustering algorithms to help train better model parameters.

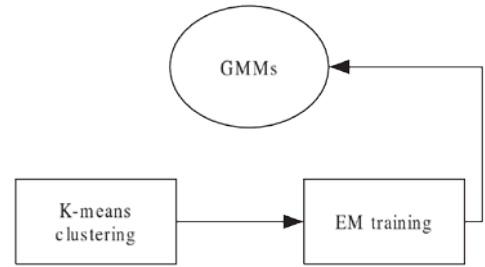

FIG. 1 THE CONVENTIONAL TRAINING ALGORITHM FOR GMMS

The ARE-TRIM is focused on modifying the first stage of the conventional GMM training algorithm, i.e., the model parameter initialisation algorithm, but keeping the fine training step with EM untouched. The modification consists of two steps: (1) substituting for the conventional K-means clustering algorithm by a more robust trimmed K-means clustering algorithm. In Section 3, we shall give more details concerning the reason why the trimmed K-means clustering algorithm is more robust than the conventional K-means clustering algorithm. (2) applying a dispersion degree model to identify the outliers automatically and then trim them off from the training population. In this procedure, no other extra step is required to govern or monitor the training process, and the overall procedure is automatic and robust and therefore is referred to as automatic robust estimation (ARE). The overall ARE-TRIM procedure is illustrated by Fig. 2.

In the next section, we shall explain why the trimmed K-means clustering algorithm is more robust than the conventional K-means clustering method and present the theorems and properties related to the trimmed K-means clustering algorithm.

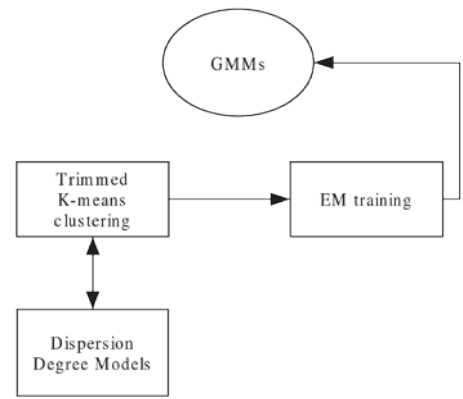

FIG. 2 THE ARE-TRIM TRAINING SCHEME FOR GMMS

---

[1] Some other model initialisation methods exist, such as mixture splitting, but the initialisation with K-means is comparable in performance to the others and also more popular. So we stick to K-means.

**Trimmed K-means clustering algorithm**

At the first step, ARE-TRIM replaces the conventional K-means clustering algorithm by the trimmed K-means clustering algorithm to initialise model parameters for GMMs. The good property of robustness of the trimmed K-means clustering algorithm is very important for improving the robustness of ARE-TRIM. Hence, in this section, we shall present some properties of the trimmed K-means clustering algorithm as well as clarify why it is more robust against gross outliers than the conventional K-means clustering algorithm.

*Algorithm Overview*

The trimmed K-means clustering is an extended version of the conventional K-means clustering algorithm by using impartial trimming scheme [3] to remove a part of data from a training population. It is based on a concept of trimmed sets. First let $D = \{\mathbf{x}_t, t \in [1, T]\}$ define a given training data set and a trimmed set is then defined by $D_\alpha$ as a subset of $D$ by trimming off $\alpha$ percent of data from the full data set $D$ in terms of a parameter $\alpha$, where $\alpha \in [0,1]$. The trimmed K-means clustering algorithm can be specifically defined as follows: the objects $\mathbf{x} \in D_\alpha$ are partitioned into $K$ clusters, $C = \{C_k\}, k = 1, \ldots, K$, based on the criter-ion of minimising intra-class covariance (MIC) $V$, i.e.,

$$C_k = \arg\min_{\mu_k} V(D_\alpha \mid C) = \arg\min_{\mu_k} \sum_{k=1}^{K} \sum_{\mathbf{x} \in D_\alpha, \mathbf{x} \in C_k} \left\| \mathbf{x} - \mu_k \right\|^2, \quad (1)$$

where $\mu_k$ is the centroid or mean point of the cluster $C_k$, which includes all the points $\mathbf{x} \in C_k$ in the trimmed set, i.e., $\mathbf{x} \in D_\alpha$ and $V(D_\alpha \mid C)$ are the intra-class covariance of the trimmed data set $D_\alpha$ with respect to the cluster set $C$, i.e.,

$$V(D_\alpha \mid C) = \sum_{k=1}^{K} \sum_{\mathbf{x} \in D_\alpha, \mathbf{x} \in C_k} \left\| \mathbf{x} - \mu_k \right\|^2. \quad (2)$$

The difference between the trimmed K-means and the conventional K-means is that the former uses a trimmed set for training, whereas the latter uses the full training popu-lation.

A solution to the above optimisation problem can be iteratively sought by using a modified Lloyd's algorithm, as follows:

**Algorithm 1** (*Trimmed K-means clustering algorithm*):

1. Initialise the centroids $\mu_k^{(0)}$ of $K$ clusters.

2. Detect outliers by following a certain principle in terms of the current centroids $\mu_k^{(n)}$, which will be described in Section 4. Meanwhile a trimmed training set $D_\alpha^{(n)}$ for the $n$-th iteration is generated.

3. Cluster all the samples in the $n$-th trimmed training set, i.e., $\mathbf{x}_t \in D_\alpha^{(n)}$ into $K$ classes, according to the following principle:

4.    $i = \arg\min_k \left\| \mathbf{x}_t - \mu_k^{(n)} \right\|^2, \forall \mathbf{x}_t \in D_\alpha$.

5. Re-calculate the centroids of $K$-clusters accord-ing to the rule of

6.    $\mu_k^{(n+1)} = \dfrac{1}{\|C_k\|} \displaystyle\sum_{\mathbf{x}_t \in D_\alpha, \mathbf{x}_t \in C_k} \mathbf{x}_t$.

7. Check the intra-class covariance $V$ as in eq. (2) to see if it is minimised [2]. If yes, stop; otherwise, go to step (1).

Existence and consistency of the trimmed K-means algorithm have been proved by Cuesta-albertos et al. in [3], where it states that, given a $p$-dimensional random vari-able space $X$, a trimming factor $\alpha$, where $\alpha \in [0,1]$, and a continuous nondecreasing metric function, there exists a trimmed K-means for $X$ [3]. Furthermore, it is also proved that for a $p$-dimensional random variable space with probability measure $P(X)$, any sequence of the em-pirical trimmed K-means converges surely in probability to the unique trimmed K-means [3].

*Why Trimmed K-means is More Robust than Conventional K-means?*

The robustness of the trimmed K-means algorithm, has been theoretically identified by using three methods in [9]: Influence Function (IF), Breakdown Point (BP) and Qualita-tive Robustness (QR). The results in [9] surely show that the trimmed K-means is theoretically more robust than the conventional K-means. Next, we shall overview these results briefly.

The IF is a metric to measure the robustness of a statis-tical estimator by providing rich quantitative and graphical information [11]. Let $x$ be a random variable and $\delta(x)$ be the probability measure, which gives mass 1 to $x$, then the IF of a functional or an estimator $T$ at a distribution $F$ is defined as the directional derivative of $T$ at $F$, in the direction of $\delta(x)$:

$$IF(x;T,F) = \lim_{t \downarrow 0}\{T((1-t)F + t\delta(x)) - T(F)\}/t \quad (5)$$

for those $x$ in which this limit exists. From the definition, we can see that the IF is a derivative of $T$ at the distribution $F$ in the direction of $\delta(x)$, as the derivative is calculated based on increasing a small amount of $T$ towards $\delta(x)$. Therefore, the IF provides a local description of the behaviour of an estimator at a probability model, such that it must always be complemented with a measure of the global reliability of the functional on the neighbourhood of the model, in order to capture an accurate view for the robustness of an estimator $T$.

---

[2] In implementation, the minimum value of the intra-class covariance can be obtained by many methods, one of which is by checking if the change of the intra-class covariance $V$ as in eq. (2) between two successive iterations is less than a given threshold.

Such a complementary measure is the so-called break-down point (BP) [5], which provides a measure of how far from the model the good properties derived from the IF's of the esimator can be expected to extend [9]. The BP measure uses the smallest fraction of corrupted observations needed to break down an estimator $T$, $\varepsilon_n^*(T, X)$, for a given data set $X$.

Besides the above two measures, the qualitative robust-ness (QR) is another method to measure the robustness of an estimator. The QR, proposed by Hampel [10] is defined via an equicontinuity condition, i.e., given a real distribution $T$ and a sequence of estimators $\{T_n\}_{n=1}^{\infty}$, we say $T$ is continu-ous if $T_n \to T(F)$ for $n \to \infty$, at a distribution $F$.

By using these three measures, we can compare the robustness between the trimmed K-means and the traditional K-means. The principal conclusions, credited to [9], can be summarised as follows:

1. IF: for the K-means, the IF is bounded only for bounded penalty (error) functions used in the clustering procedure, whereas it is bounded for the trimmed K-means for a wider range of functions and practically IF vanishes outside each cluster.

2. BP: the smallest fraction of corrupted observations needed to break down the K-means estimator, $\varepsilon_n^*(T, X)$, is $1/n$ for a given data set $X$ with $n$ points, where it could be near to 1 over the trimming size for well clustered data for the trimmed K-means. Thus, the BP of the trimmed K-means is much larger.

3. QR: There is no QR for the K-means in the case of $K \geq 2$ while QR exists for every $K \geq 1$ for the trimmed K-means.

The theory of robust statistics has laid solid theoretic foundations for the robustness of the trimmed K-means, which in turns provide strong supports for the advantages of the proposed ARE-TRIM over the conventional GMM training algorithm.

The robustness of the trimmed K-means can be illustrated by Fig. 3. In Fig. 3, two clusters are assumed to represent two groups of data $A$ and $B$. Most of data $A$ and $B$ are clustered around their clusters $C_1$ and $C_2$, except an outlier $A'$ for the cluster $C_1$. In fact, the outlier point $A'$ is referred to as a breakdown point, i.e., BP. In this case, the classic K-means surely breaks down due to the existence of the BP $A'$, and therefore two clusters $C_1'$ and $C_2'$ are generated. However, for the trimmed K-means, the algorithm does not break down with an appropriate trimm-ing value, and the two right clusters $C_1$ and $C_2$ are still able to be sought. This illustrates the robustness of the trimmed K-means.

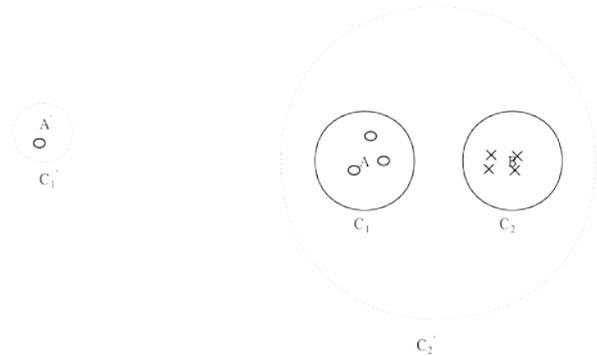

FIG. 3 ILLUSTRATION TO THE ROBUSTNESS OF THE TRIMMED K-MEANS CLUSTERING

## Modelling data dispersion degree for automatic robust estimation

In previous work [3], a fixed fraction of data is trimmed off from a full training set, in order to realise the training of the trimmed K-means. However, in real applications, e.g., acoustic noise reduction and speaker recognition tasks, a fixed cut-off data strategy may not be able to successfully trim off all the essential outliers, as test data do not often have a fixed number of outliers. In this paper, we propose that this issue can be solved by using a model of data dispersion degree.

We start our definition for data dispersion degree from the simplest case, i.e.,the dispersion degree of a data point in terms of a cluster. For a cluster $c$ with a mean vector $\mu$ and an inverse covariance matrix $\Sigma^{-1}$, a dispersion degree, $d(\mathbf{x}||c)$, of a given data point $\mathbf{x}$, with which the data point deviates from the cluster $c$ is defined as a certain distance

$$d(\mathbf{x}||c) = distance(\mathbf{x}||c). \qquad (6)$$

A variety of distances can be adopted for this metric. For instance, the Euclidean distance can be used to represent the dispersion degree of a data point $\mathbf{x}$ and a cluster $c = \{\mu, \Sigma^{-1}\}$, i.e.,

$$d(\mathbf{x}||c) = \|\mathbf{x} - \mu\| = \sqrt{\sum_{i=1}^{d}(x_i - \mu_i)^2}, \qquad (7)$$

where $d$ is the dimension of the vector $\mathbf{x}$.

Apart from this, the Mahalanobis distance (M-distance) can be also used. By this metric, the dispersion degree of a data point $\mathbf{x}$ in terms of the class $c = \{\mu, \Sigma^{-1}\}$ can be defined as follows:

$$d(\mathbf{x}||c) = (\mathbf{x} - \mu)^T \Sigma^{-1}(\mathbf{x} - \mu). \qquad (8)$$

The concept of dispersion degree of a data point in terms of one cluster can be easily generalised into a multi-class case. For a multi-class set $\mathbf{C} = \{c_k\}, k \in [1, K]$, the dis-persion degree $d(\mathbf{x}||\mathbf{C})$ of a data point $\mathbf{x}$ in terms of a multi-class set $\mathbf{C}$ is then defined as the dispersion degree of the data point $\mathbf{x}$ in terms of the optimal class to which the data point belongs in a sense of the Bayesian decision rule, i.e.,

$$d(\mathbf{x}||\mathbf{C}) = d(\mathbf{x}||c_j), \qquad (9)$$

where

$$j = \arg\max_{k=1}^{K} Pr(\mathbf{x}|c_k) \qquad (10)$$

and $Pr(\mathbf{x}|c_k)$ is a conditional probability of the data $\mathbf{x}$ generated by $c_k$. In statistical modelling theory, the condi-tional probability $Pr(\mathbf{x}|c_k)$ often takes a form of normal distribution with a mean vector $\mu_k$ and an inverse cova-riance matrix $\Sigma_k^{-1}$, e.g.,

$$Pr(\mathbf{x}|c_k) = \mathsf{N}(\mathbf{x}|\mu_k, \Sigma_k^{-1}). \qquad (11)$$

Based on the above definitions, we can further define the dispersion degree of a data set $\mathbf{X} = \{\mathbf{x}_t\}, t \in [1, T]$

in terms of a multi-class $\mathbf{C} = \{c_k\}, k \in [1, K]$ as a sequence of the dispersion degree of each data point $\mathbf{x}_t$ in terms of a multi-class set $\mathbf{C}$, i.e.,

$$d(\mathbf{X}||\mathbf{C}) = \{d(\mathbf{x}_t||\mathbf{C})\}, t \in [1, T].$$

Given a data set $\mathbf{X}$, if we use a random variable $y$ to represent the dispersion degree of a data set $\mathbf{X}$ in terms of a multi-class set $\mathbf{C}$, then we can prove that under a certain distance metric, the random variable $y$ approximately ob-eys a certain distribution when $T$, the size of the data set $X$, approaches infinity, i.e., $T \to +\infty$. For this, we have our first theorem with respect to the application of the Euclidean distance as below.

**Theorem 4.1** *Let the random variable* $\mathbf{X} = \{\mathbf{x}_1, \mathbf{x}_2, \cdots, \mathbf{x}_T\}$ *be a set of data points to represent a d-dimensional space and can be optimally modelled by a GMM $G$, i.e.,*

$$\mathbf{X} \sim G = \sum_{k=1}^{K} w_k \mathsf{N}(\mu_k, \Sigma_k^{-1}), \qquad (12)$$

*Where $K$ is the number of Gaussian components of the GMM $G$, $w_k$, $\mu_k$ and $\Sigma_k^{-1}$ are the weight, mean and dia-gonal inverse covaraince matrix of the $k$-th Gaussian with $\sigma_{ki}^{-2}, 1 \le i \le d$ at the diagonal of $\Sigma_k^{-1}$.*

*Let $\mathbf{C} = \{c_k\}, k \in [1, K]$ denote a multi-class set, wh-ere each of its elements $c_k$ represents a Gaussian compo-nent $\mathsf{N}(\mu_k, \Sigma_k^{-1})$ for the GMM $G$. Each sample $\mathbf{x}_t$ of the data set $\mathbf{X}$ can be assigned to an optimal class $c_j$, under the Bayesian rule, i.e., selecting the maximum conditional probability of the data point $\mathbf{x}_t$ given a class $c_k$, which is the k-th Gaussian of the model $G$,*

$$j = \arg\max_{k=1}^{K} Pr(\mathbf{x}_t|c_k) = \arg\max_{k=1}^{K} \mathsf{N}(\mathbf{x}_t|\mu_k, \Sigma_k^{-1}). \qquad (13)$$

*Let's also assume a random variable $y$ to represent the dispersion degree of the data set $\mathbf{X}$ in terms of the multi-class $\mathbf{C}$, which includes the dispersion degree of each sample in the data set $\mathbf{X}$ that is derived based on the conti-nuous Euclidean distance as defined in eq. (7), i.e.,*

$$y = \{\|\mathbf{x}_t^{(k)} - \mu_k\|\}, k \in [1, K], \text{and } t \in [1, T]. \qquad (14)$$

*Then we can prove that the distribution of the random variable $y$ is approximately a chi distribution $\chi(x; \nu)$ with the $\nu$ degrees of freedom ([1]), i.e.,*

$$y \sim \chi(x; \nu) = \frac{2^{1-\frac{\nu}{2}} x^{\nu-1} e^{-\frac{x^2}{2}}}{\Gamma(\frac{\nu}{2})}, \qquad (15)$$

*where*

$$\nu = \sum_{i=1}^{K} \frac{(\sum_{j=1}^{d} \sigma_{ij}^4)^3}{(\sum_{j=1}^{d} \sigma_{ij}^6)^2} \qquad (16)$$

*and* $\Gamma(z)$ *is the Gamma function,*

$$\Gamma(z) = \int_0^\infty t^{z-1} e^{-t} dt. \qquad (17)$$

**Proof**: By applying the decision rule based on the optimal conditional probability distribution of a data point $\mathbf{x}_t$ and a class $c_k$, we can partition the overall data set $\mathbf{X}$ into a disjoint set, i.e.,

$$\mathbf{X} = \{\mathbf{x}^{(1)}, \mathbf{x}^{(2)}, \cdots, \mathbf{x}^{(K)}\}. \qquad (18)$$

According to the essential training of GMM, e.g., the ML training, any data point $\mathbf{x}_t$ is *assumed* to be modelled by a normal distribution component of GMM, thus, we naturally have $\mathbf{x}^{(k)} \sim \mathsf{N}(\mu_k, \Sigma_k^{-1})$ in a sense of the Bayesian decision rule. Thus we have

$$\mathbf{x}^{(k)} - \mu_k \sim \mathsf{N}(0, \Sigma_k^{-1}).$$

According to eq. (7), we know that

$$y_{(k)}^2 = \left\| \mathbf{x}^{(k)} - \mu_k \right\|^2 = \sum_{i=1}^{d} \sigma_{ki}^2 \left(\frac{x_i^{(k)} - \mu_{ki}}{\sigma_{ki}}\right)^2 = \sum_{i=1}^{d} \sigma_{ki}^2 \cdot \chi^2(1), \qquad (19)$$

thus, $y_{(k)}^2$ is in fact a linear combination of chi-square dis-tributions $\chi^2(1)$ with 1 degree of freedom.

According to [13], it is easy to verify that $y_{(k)}^2$ approximately obeys a chi-square $\chi^2(\nu_k)$, where

$$\nu_k = \frac{(\sum_{i=1}^{d} \sigma_{ki}^4)^3}{(\sum_{i=1}^{d} \sigma_{ki}^6)^2} \qquad (20)$$

(see Appendix for detailed derivation).

To this point, we should be able to get a sequence of variables $y_{(k)}^2$, each of which is a $\chi^2(\nu_k)$ distribution with $\nu_k$ degrees of freedom, for each partition $\mathbf{x}^{(k)}$ of the data $X$.

If we define a new variable $y$, the square of which can be expressed as follows:

$$y^2 = \frac{\sum_{k=1}^{K} y_{(k)}^2}{K}, \qquad (21)$$

then we can prove that $y^2$ approximately obeys another chi-square distribution $\chi_{y^2}^2(\nu)$, with the degrees of free-dom $\nu$, where

$$\nu = \sum_{i=1}^{K} \frac{(\sum_{j=1}^{d} \sigma_{ij}^4)^3}{(\sum_{j=1}^{d} \sigma_{ij}^6)^2}. \qquad (22)$$

The proof is straightforward by noticing that $y^2$ is in fact a linear combination of chi-square distributions $y_{(k)}^2$. Thus, we can prove this theorem by using Theorem 6.1. For this, we can have the parameters as follows:

$$\delta_k = 0 \qquad (23)$$

$$h_k = \nu_k = \frac{(\sum_{i=1}^{d} \sigma_{ki}^4)^3}{(\sum_{i=1}^{d} \sigma_{ki}^6)^2} \qquad (24)$$

$$\lambda_k = \frac{1}{K} \qquad (25)$$

$$c_k = \sum_{i=1}^{K} \frac{1}{K^k} \frac{(\sum_{j=1}^{d} \sigma_{ij}^4)^3}{(\sum_{j=1}^{d} \sigma_{ij}^6)^2} \qquad (26)$$

$$s_1 = \frac{c_3}{c_2^{3/2}} \qquad (27)$$

$$s_2 = \frac{c_4}{c_2^2}. \qquad (28)$$

Because

$$\frac{s_1^2}{s_2} = \frac{c_3^2}{c_2 c_4} = \frac{\frac{1}{K^6} p^2}{\frac{1}{K^2} p \cdot \frac{1}{K^4} p} = 1, \qquad (29)$$

where

$$p = \sum_{i=1}^{K} \frac{(\sum_{j=1}^{d} \sigma_{ij}^4)^3}{(\sum_{j=1}^{d} \sigma_{ij}^6)^2},\qquad(30)$$

we have $s_1^2 = s_2$. Therefore, we know that $y^2$ is approxi-mately a central chi-square distribution $\chi_y^2(\nu, \delta)$ with

$$\delta = 0 \qquad (31)$$

$$\nu = \frac{c_2^3}{c_3^2} = \frac{\frac{1}{K^6} p^3}{\frac{1}{K^6} p^2} = p. \qquad (32)$$

Further, we can naturally obtain that $y \sim \chi_y(\nu)$. Then the proof for this theorem is done.

*Remarks*: Theorem 4.1 shows that the dispersion degree of a data set in terms of the Euclidean distance approxi-mately obeys a chi distribution. However, when the degrees of freedom $\nu$ is large, a $\chi$ distribution can be approxi-mated by a normal distribution $\mathsf{N}(\mu, \sigma)$. This result is es-pecially useful for real applications, for its easy computation, when $\nu$ is very large. To this end, the following theorem holds [2]:

**Theorem 4.2** *A chi distribution $\chi(\nu)$ as in eq. (15) can be accurately approximated by a normal distribution $\mathsf{N}(\mu, \sigma^2)$ for large $\nu$ s with $\mu = \sqrt{\nu - 1}$ and $\sigma^2 = 1/2$ [2].*

**Proof**: The proof is straightforward by using Laplace appro-ximation [2, 21]. According to Laplace approximation, a given pdf $p(x)$ with its log-pdf $f(x) = \ln p(x)$ can be approximated by a normal distribution $\mathsf{N}(x_{max}, \sigma^2)$, where $x_{max}$ is a local maximum of the log-pdf $f(x)$ and $\sigma^2 = -\frac{1}{f''(x_{max})}$.

So the proof is to find $x_{max}$ and $\sigma^2$. For this, we can derive

$$f(x) = \ln p(x) = \ln x^{\nu-1} + \ln e^{-\frac{x^2}{2}} \qquad (33)$$

by using the definition of chi distribution as in eq. (15) and ignoring the constant $\Gamma(\frac{\nu}{2})$ and $2^{1-\frac{\nu}{2}}$.

By setting the first-order derivative of $f(x)$ to zero, we get

$$\frac{df(x)}{dx} = \frac{\nu - 1}{x} - x = 0. \qquad (34)$$

Hence, $x_{max} = \sqrt{\nu - 1}$ and

$$\sigma^2 = -\frac{1}{f''(x) \mid x_{max}} = -1\frac{1}{\frac{d(\frac{\nu-1}{x} - x)}{dx} \mid x_{max}} = \frac{1}{2}. \qquad (35)$$

The derivation is done.

Next, we shall present the other main result of dispersion degree modelling in regard to using the Mahalanobis dis-tance in the following theorem.

**Theorem 4.3** *Let the random variable $\mathbf{X} = \{\mathbf{x}_1, \mathbf{x}_2, \cdots, \mathbf{x}_T\}$ be a set of data points to represent a d-dimensional space and can be optimally modelled by a GMM $G$, i.e.,*

$$\mathbf{X} \sim G = \sum_{k=1}^{K} w_k \mathsf{N}(\mu_k, \Sigma_k^{-1}), \qquad (36)$$

*where $K$ is the number of Gaussian components of the GMM $G$, $w_k$, $\mu_k$ and $\Sigma_k^{-1}$ are the weight, mean and dia-gonal inverse covaraince matrix of the $k$-th Gaussian with $\sigma_{ki}^{-2}, 1 \le i \le d$ at the diagonal of $\Sigma_k^{-1}$.*

Let $\mathbf{C} = \{c_k\}, k \in [1, K]$ denote a multi-class set, where each of its elements $c_k$ represents a Gaussian com-ponent $\mathsf{N}(\mu_k, \Sigma_k^{-1})$ for the GMM $G$. Each sample $\mathbf{x}_t$ of the data set $\mathbf{X}$ can be assigned to an optimal class $c_j$, under the Bayesian rule, i.e., selecting the maximum conditional probability of the data point $\mathbf{x}_t$ given a class $c_k$, which is the k-th Gaussian of the model $G$,

$$j = \arg\max_{k=1}^{K} Pr(\mathbf{x}_t \mid c_k) = \arg\max_{k=1}^{K} \mathsf{N}(\mathbf{x}_t \mid \mu_k, \Sigma_k^{-1}). \qquad (37)$$

Let's also assume a random variable $y$ to represent the dispersion degree of the data set $\mathbf{X}$ in terms of the multi-class $\mathbf{C}$, which includes the dispersion degree of each sample in the data set $\mathbf{X}$ that is derived based on the continuous Mahalanobis distance as defined in eq. (8), i.e.,

$$y = \{(\mathbf{x}_t^{(k)} - \mu_k)^T \Sigma_k^{-1}(\mathbf{x}_t^{(k)} - \mu_k) : k \in [1, K], t \in [1, T]\}. \qquad (38)$$

*Then the distribution of the random variable* $y$ *is approximately a chi-square distribution with* $\nu$ *degrees of freedom [1], i.e.,*

$$y \sim \chi^2(x;\nu) = \frac{1}{2^{\nu/2}\,\Gamma(\nu/2)}\, x^{\nu/2-1} e^{-x/2} \qquad (39)$$

*where*

$$\nu = dK. \qquad (40)$$

**Proof**: Following the same strategy, the data set $X$ can also be partitioned into a class $X = \{\mathbf{x}^{(1)}, \mathbf{x}^{(2)}, \cdots, \mathbf{x}^{(K)}\}$. By noting that

$y_{(k)} = (\mathbf{x}^{(k)} - \mu_k)^T \Sigma_k^{-1}(\mathbf{x}^{(k)} - \mu_k) \sim \chi^2(x^{(k)}, d)$, then we get $y_{(k)} \sim \chi^2(x^{(k)}, d)$. Further, let

$$y = \frac{\sum_{k=1}^{K} y_{(k)}}{K}, \qquad (41)$$

then we can prove that

$$y \sim \chi^2(x, \nu) \qquad (42)$$

where $\nu$ is given by eq. (40).

The prove is a straightforward result by applying Theo-rem 6.1, as $y$ is in fact a linear combination of $\chi^2$ distri-butions of $y_{(k)}$. For this, we can easily know that

$$\lambda_k = \frac{1}{K} \qquad (43)$$

$$h_k = d \qquad (44)$$

$$\delta_k = 0 \qquad (45)$$

$$c_k = \sum_{i=1}^{K}(\frac{1}{K})^k d = \frac{d}{K^{k-1}} \qquad (46)$$

$$s_1^2 = \frac{c_3^2}{c_2^3} = \frac{1}{dK} \qquad (47)$$

$$s_2 = \frac{c_4}{c_2^2} = \frac{1}{dK}. \qquad (48)$$

As $s_1^2 = s_2$, therefore, $y$ approximately obeys a central $\chi^2$ distribution $\chi^2(\nu, \delta)$ with $\delta = 0$ and

$$\nu = \frac{c_2^3}{c_3^2} = dK\,.$$

Then the proof is done.

From these two theorems, we can see that the dispersion degree of a data set in terms of a multi-class set can be therefore modelled by either a chi distribution or a chi-square distribution depending on the distance measure applied to it. These are the theoretical results. In practice, the normal distribution is often used instead due to its fast computation and convenient manipulation, to approximate a chi and chi-square distribution, especially when the number of the degrees of freedom of the chi (chi-square) distribution is large, e.g., $\nu > 10$. For the $\chi$ distribution in eq. (15), we have known that it can be approximated by using Theorem 4.2. While for the $\chi^2$ distribution in eq. (39), we have the following result.

**Theorem 4.4** *A chi-square distribution* $\chi^2(\nu)$ *can be approximated by a normal distribution* $\mathsf{N}(\nu, 2\nu)$ *for large* $\nu$ *s, where* $\nu$ *is the number of the degree of freedom of* $\chi^2(\nu)$ *[12].*

After clarifying the dispersion degree essentially obeys a certain distribution, we can apply this model to automatic outlier removal in ARE-TRIM by using the following definition:

**Definition 4.1** *Given a dispersion degree model* $\mathsf{M}$ *for a data set* $\mathbf{X}$ *and a threshold* $\tau$, *where* $\tau > 0$, *any data point* $\mathbf{x}$ *is identified as an outlier at the threshold* $\tau$, *if the conditional cumulative probability* $P(\mathbf{x}|\mathsf{M})$ *of the dispersion degree of the data point* $\mathbf{x}$ *conditioned on the dispersion degree model* $\mathsf{M}$ *is larger than a threshold* $\tau$, *i.e.,*

$$P(\mathbf{x}|\mathsf{M}) = \int_{-\infty}^{d(\mathbf{x}\mathbf{PC})} p(y|\mathsf{M})dy > \tau. \qquad (49)$$

With Definition 4.1 and a proper value selected for the threshold $\tau$, the outliers can be automatically identified and thus removed by our proposed ARE-TRIM training scheme. The detailed algorithm is formulated as follows, including two main parts, i.e., the estimation and identification pro-cesses:

**Algorithm 2 (ARE-TRIM training algorithm)** :

1. Estimation of the model $M$ : according to Theo-rem 4.1, 4.3, 4.2, 4.4, $y \sim \mathsf{N}(\theta)$, where $\theta = (\mu, \sigma^2)$. Then $\theta$ can be asymptotically estimated as $\hat{\theta} = (\hat{\mu}, \hat{\sigma}^2)$ by the well-known formulae

$$\hat{\mu} = \frac{\sum_{t=1}^{T} y_t}{T} \qquad (50)$$

$$\hat{\sigma} = \sqrt{\frac{1}{T-1}(\sum_{t=1}^{T}(y_t - \hat{\mu})^2)} \qquad (51)$$

where

$$y_t = d(\mathbf{x}_t \parallel \mathbf{C}). \qquad (52)$$

2. Outlier identification:

(a) For each data sample $\mathbf{x}_t$ do;

i. Calculate $y_t$ according to eq. (52);

ii. Calculate the cumulative probability $P(\mathbf{x}_t \mid \mathsf{M})$ according to eq. (49);

iii. If $P(\mathbf{x}_t \mid \mathsf{M}) > \tau$, then $\mathbf{x}_t$ is identified as an outlier and is trimmed off from the training set; otherwise $\mathbf{x}_t$ is not an outlier and thus used for GMM training.

(b) Endfor

*Further Remarks*: In the theory, you may notice that both of theorem 4.1 and 4.3 reply on using the Bayesian decision rule for the partition process. As it is well known in pattern recognition domain [6], the Bayesian decision rule is an optional classifier in a sense of minimising the decision risk in a squared form [6]. Hence, the quality of the recog-nition process should be able to satisfy the requirements of most of the pattern recognition applications. In the next section, we shall carry out experiments to show that such a procedure is effective.

## Experiments

In previous sections, we have discussed the ARE-TRIM algorithm from a theoretical viewpoint. In this section, we shall show its effectiveness by applying it to a real signal processing application.

Our proposed training approach aims at improving the robustness of the classic GMMs by adopting the automatic trimmed K-means training techniques. Thus, theoretically speaking, any application using GMMs, such as speaker /speech recognition or acoustical noise reduction, can be used to evaluate the effectiveness of our proposed method. Without loss of genera-lisation, we select a moderately-sized speaker recognition task for evaluation, as GMM is widely accepted as the most effective method for speaker recognition.

Speaker recognition has two common application tasks: speaker identification (SI) (recognising a speaker identity) and speaker verification (SV) (authenticating a registered valid speaker) [17]. Since the classic GMMs have been demonstrated to be very efficient for SI [19], we simply choose the SI task based on a telephony corpus - NTIMIT [7] - for evaluation.

In NTIMIT, there are ten sentences (5 SX's, 2 SA's and 3 SI's) for each speaker. Similar to [16], we used six sentences (two $SX_{1-2}$, two $SA_{1-2}$ and two $SI_{1-2}$) as train-ing set, two sentences ($SX_3$ and $SI_3$) as development set and the last two SX utterances ($SX_{4-5}$) as test set. The development set is used for fine tuning the relevant para-meters for GMM training. With it, we select a variance threshold factor of 0.01 and minimum Gaussian weight of 0.05 as optimum values for GMM training (performance falling sharply if either is halved or doubled).

As in [14, 17, 26-28], MFCC features, obtained using HTK [29], are used, with 20ms windows and 10ms shift, a pre-emphasis factor of 0.97, a Hamming window and 20 Mel scaled feature bands. All 20 MFCC coefficients are used except c0. On this database, neither cepstral mean subtraction, nor time difference features increase perform-ance, so these are not used. Apart from these, no extra processing measure is employed.

Also as in [16-18], GMMs with 32 Gaussians are trained to model each speaker for SI tasks. All the Gaussians use diagonal covariance matrices, as it is well-known in speech domain that diagonal converiance matrices produce very similar results to full converiance matrices [15, 29]. Also, the standard MLE method [2] based on the EM algorithm [4] is used to train GMMs, due to its efficiency and wide appli-cation in speech processing.

In the trimmed K-means step, random selection of $K$ points is used to initialise the centroids of the $K$ clusters.

### Experiments with Euclidean distance

We first present the experiment of testing the dispersion degree model by using the Euclidean distance. In this expe-riment, different values for the threshold $\tau$, according to Definition 4.1, from 1.0 to 0.6 are used to trim off the outliers existing in the training data. $\tau = 1.0$ represents no outlier is pruned, whereas $\tau = 0.6$ means a maximum number of outliers are identified and trimmed off. The proposed method is then tested on both the development set and test set. The development test is used to select the optimal value for the trimming threshold $\tau$. The results are presented in Fig. 4.

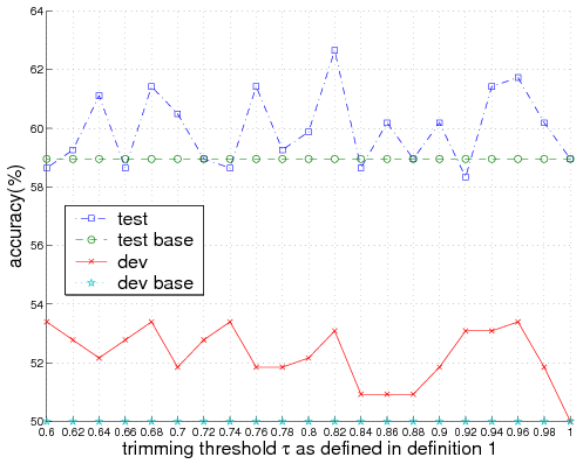

FIG. 4 THE ARE-TRIM GMMS VS. THE CONVENTIONAL GMMS ON NTIMIT WITH THE EUCLIDEAN DISTANCE

From Fig. 4, we can see that: (1) the proposed method does improve system performance on both the development and test set with the threshold $\tau \in [0.6, 1.0)$. The accuracy of ARE-TRIM for all the threshold values on the develop-ment and for most of them on the test set is higher than that of the conventional GMM training method. This shows the effectiveness of the proposed method. (2) The values of threshold $\tau$ can not be too small; Otherwise, they will remove too much meaningful points that are actually not outliers. It may result in unpredictable system performance (though most of them are still helpful to improve system performance as shown in Fig. 4). This can be showed in Fig. 5, where we give the averaged proportions of the trimmed samples corresponding to each value of the threshold $\tau$.

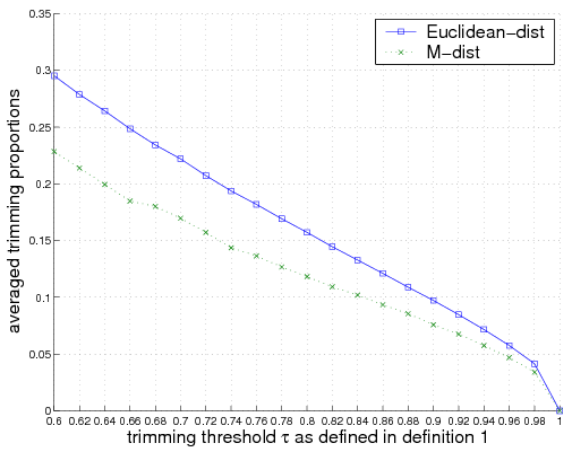

FIG. 5 THE AVERAGED PROPORTIONS OF TRIMMING DATA REGARDING DIFFERENT THRESHOLD VALUES BASED ON THE EUCLIDEAN DISTANCE (EUCLIDEAN-DIST) AND THE MAHALANOBIS DISTANCE (M-DIST)

The "averaged" proportions are obtained across from a number of training speakers and a number of iterations. Before each iteration of the EM training, a trimmed K-means is applied using a trimmed set with the threshold $\tau$. From this figure, we can see that the averaged trimming proportions increase when the values of the trimming threshold $\tau$ move away from $1.0$. (3) In practice, we suggest the range of $[0.9, 1.0]$ be used to select an optimal value for $\tau$, as it is a reasonable probability range for identifying outliers. Out of this range, valid data are highly possibly trimmed off. This can be also partly shown in Fig. 5, as $0.9$ is roughly corresponding to $10\%$ of data being removed. (4) The trend of performance changes on the development and test set shows a similar increase before the peak values are obtained at the $6\%$ outlier removal, with the improvements from $50.0\%$ to $53.40\%$ on the development set and from $58.95\%$ to $61.73\%$ on the test set. After $6\%$ outlier re-moval, system performance varies on both the development and test set. It implies that the outliers in the training data have been effectively removed and more robust models are obtained. However, when more data are removed with $\tau$ taking values beyond $0.96$, useful data are removed as well. Hence, system performance is demonstrated as a threshold-varying characteristic, depending on which part of data is removed. Therefore, we suggest in practice $[0.90, 1.0)$ be used, from which a suitable value for the threshold $\tau$ is selected. In this experiment, we choose $0.96$ as an optimum for $\tau$ based on the development set.

*Experiments with the Mahalanobis distance*

When we use the Euclidean distance to model dispersion degree, we do not take into consideration the covariance of the data in a cluster but only consider the distances to the centroid. However, the covariance of a cluster may be quite different, and it may largely affect the distribution of data dispersion degrees. Thus, in this experiment, we evaluate the use of the Mahalanobis distance for the automatic trim-ming measure in ARE-TRIM.

From Fig. 6, we can see that for both the development and test set, the trimmed training scheme can significantly improve the robustness and system performance. The highest accuracy on the development set is improved from $50.0\%$ to $55.25\%$, with the trimming factor $\tau = 0.92$, where the accuracy for the test set is improved from $58.95\%$ to $61.11\%$. This shows that the automatic trimming scheme, i.e., ARE-TRIM, is quite effective to improve the robustness of GMMs.

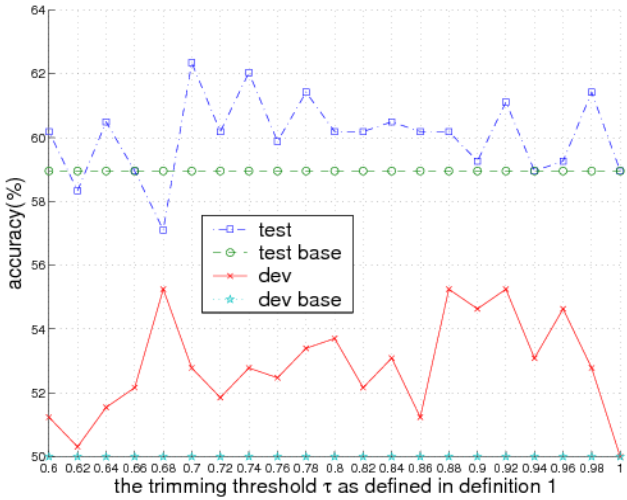

FIG. 6 THE ARE-TRIM GMMS VS. THE CONVENTIONAL GMMS ON NTIMIT WITH THE MAHALANOBIS DISTANCE

When the values of the trimming threshold are too large, similar to the trimming measure with the Euclidean distance, speaker identification accuracy shows a certain threshold-varying behaviour, because, in this case, not only the out-liers but also some meaningful data samples are trimmed off as well. This is similar to the case with the Euclidean distance. Furthermore, by comparing Fig. 4 and 6, we can find that the improvements on the development set by using the Mahalanobis distance ( 55.25% ) are larger than those obtained by using the Euclidean distance ( 53.39% ). It may suggest that the Mahalanobis distance is a better metric to model dispersion degrees, because of consideration of the covariance factor in modelling.

## Conclusions

In this paper, we have proposed an automatic trimming estimation algorithm for the conventional Gaussian mixture models. This trimming scheme consists of several novel contributions to improve the robustness of Gaussian mixture model training by effectively removing outlier interference. First of all, a modified Lloyd's algorithm is proposed to realise the trimmed K-means clustering algorithm and used for parameter initialisation for Gaussian mixture models. Secondly, data dispersion degree is proposed to be used for automatically identifying outliers. Thirdly, we have theore-tically proved that data dispersion degree in the context of GMM training approximately obeys a certain distribution (chi and chi-square distribution), in terms of either the Eu-clidean or

Mahalanobis distance being applied. Finally, the proposed training scheme has been evaluated on a realistic application with a medium-size speaker identification task. The experiments have showed that the proposed method can significantly improve the robustness of Gaussian mixture models.

## Appendix

### Theorem from [13]

We shall first cite the result from [13] as Theorem 6.1, and then use it to derive the result used in the proof for Theorem 4.1.

**Theorem 6.1** *Let $Q(X)$ be a weighted sum of non-central chi-square variables, i.e.,*

$$Q(X) = \sum_{i=1}^{m} \lambda_i \chi_{h_i}^2(\delta_i), \tag{53}$$

where $h_i$ is the degrees of freedom and $\delta_i$ is the non-centrality parameter of the $i$-th $\chi^2$ distribution. Define the following parameters

$$c_k = \sum_{i=1}^{m} \lambda_i^k h_i + k \sum_{i=1}^{m} \lambda_i^k \delta_i \tag{54}$$

$$s_1 = c_3 / c^{3/2} \tag{55}$$

$$s_2 = c_4 / c_2^2, \tag{56}$$

then $Q(X)$ can be approximated by a chi-square distri-bution $\chi_l^2(\delta)$, where the degrees of freedom $l$ and the non-centrality $\delta$ are divided into two cases:

$$l = \begin{cases} c_2^3/c_3^2, & \text{if } s_1^2 \le s_2 \\ a^2 - 2\delta, & \text{if } s_1^2 > s_2, \end{cases} \tag{57}$$

$$\delta = \begin{cases} 0, & \text{if } s_1^2 \le s_2 \\ s_1 a^3 - a^2, & \text{if } s_1^2 > s_2 \end{cases} \tag{58}$$

and

$$a = 1/(s_1 - \sqrt{s_1^2 - s_2}). \tag{59}$$

### Useful result for the proof of Theorem 4.1

Next, we shall use Theorem 6.1 to derive the distri-bution of $y_{(k)}^2$ in eq. (19) used in the proof of Theorem 4.1. For the simplicity of presentation, we drop off the superscript $(k)$ without any confusion, as it is clear in the context that this derivation procedure is regarding the $k$-th partition of the data set $X$. From eq. (19), we know that $y^2$ is a linear combination of $1$-dimensional $\chi^2$ distri-bution, i.e.,

$$y^2 = \sum_{i=1}^{d} \sigma_i^2 \cdot \chi^2(1). \tag{60}$$

As $\chi^2(1)$ is central, we have $h_i = 1$, $\delta_i = 0$ and $\lambda_i = \sigma_i^2$ in our case. With these quantities, it is easy to know that

$$c_k = \sum_{i=1}^{d} \sigma_i^{2k} \tag{61}$$

$$s_1 = \frac{c_3}{c_2^{3/2}} = \frac{\sum_{i=1}^{d}\sigma_i^6}{(\sum_{i=1}^{d}\sigma_i^4)^{\frac{3}{2}}} \tag{62}$$

$$s_2 = \frac{c_4}{c_2^2} = \frac{\sum_{i=1}^{d}\sigma_i^8}{(\sum_{i=1}^{d}\sigma_i^4)^2}. \tag{63}$$

From this we can know

$$s_1^2 \le s_2, \tag{64}$$

because

$$\frac{s_1^2}{s_2} = \frac{(\sum_{i=1}^{d}\sigma_i^6)^2}{\sum_{i=1}^{d}\sigma_i^4 \cdot \sum_{i=1}^{d}\sigma_i^8} = \frac{\sum_{i=1}^{d}\sigma_i^{12} + \sum_{i}\sum_{j,j\neq i}\sigma_i^6\sigma_j^6}{\sum_{i=1}^{d}\sigma_i^{12} + \sum_{i}\sum_{j,j\neq i}\sigma_i^4\sigma_j^8} \le 1, \tag{65}$$

and

$$C = \sum_{i}\sum_{j,j\neq i}\sigma_i^6\sigma_j^6 - \sum_{i}\sum_{j,j\neq i}\sigma_i^4\sigma_j^8 = \sum_{i}\sum_{j,j\neq i}\sigma_i^4\sigma_j^6(\sigma_i^2 - \sigma_j^2). \tag{66}$$

By noticing a term $A$

$$A = \sigma_i^4\sigma_j^6(\sigma_i^2 - \sigma_j^2) \tag{67}$$

always has a symmetric pair, a term $B$, i.e.,

$$B = \sigma_j^4\sigma_i^6(\sigma_j^2 - \sigma_i^2), \tag{68}$$

then we can reorganise them as

$$A + B = \sigma_i^4\sigma_j^4(\sigma_i^2 - \sigma_j^2)(\sigma_j^2 - \sigma_i^2) \le 0. \tag{69}$$

Thus, we know $C \le 0$ and further

$$\sum_{i}\sum_{j,j\neq i}\sigma_i^6\sigma_j^6 \le \sum_{i}\sum_{j,j\neq i}\sigma_i^4\sigma_j^8. \tag{70}$$

Therefore we have eq. (64).

As we know $s_1^2 \le s_2$, by using Theorem 6.1, we can obtain that the approximated $\chi_i^2(\delta)$ is central with the parameters

$$\delta = 0 \tag{71}$$

$$l = \frac{c_2^3}{c_3^2} = \frac{(\sum_{i=1}^{d}\sigma_i^4)^3}{(\sum_{i=1}^{d}\sigma_i^6)^2}. \tag{72}$$

## REFERENCES

Alexander, M, Graybill, F., & Boes, D., Introduction to the Theory of Statistics (Third Edition), *McGraw-Hill*, ISBN 0-07-042864-6, 1974.

Bishop, C. M., *Pattern recognition and machine learning*, Springer Press, 2006.

Cuesta-Albertos, J. A., Gordaliza, A., & Matran, C., "Trimmed K-means: a attempt to robustify quantizers", *The Annals of Statistics*, 25, 2, 553 – 576, 1997.

Dempster, A. P., Laird, N. M., & Rubin, D. B., "Maximum likelihood from incomplete data via the EM algorithm (with discussion)", *Journal of the Royal Statistical Society B*, 39, 1 – 38, 1977.

Donoho, D. L., & Huber, P. J., The notion of breakdown point. *A Festschrift for Erich L. Lehmann*, eds. P. Bickel, K. Doksum and J. L. Hodges, Jr. Belmont, CA: Wadsworth, 1983.

Duda, O., Hart, P.E., & Stork, D.G., *Pattern classification*, Wiley Press, 2001.

Fisher, W. M., Doddington, G. R., & Goudie-Marshall, K. M., "The DARPA speech recognition research database: Specifications and status", *Proceedings of DARPA Workshop on Speech Recognition*, 93 – 99, February, 1986.

Forsyth, D. A., & Ponce, J., Computer Vision: A Modern Approach, Prentice Hall, 2002.

Garcia-Escudero, L. A., & Gordaliza, A., "Robustness properties of K-means and trimmed k-means", *Journal of the American Statistical Association*, 94, 956 – 969, 1999.

Hampel, F.R., "A general qualitative definition of robustness", *Annals of Mathematical Statistics*, 42, 1887 – 1896, 1971.

Hampel, F. R., Rousseeuw, P.J., Ronchetti, E., & Stahel, W. A., *Robust Statistics: The Approach based on the Influence Function*, New York, Wiley, 1986.

I. Shafran and R. Rose, "Robust speech detection and segmentation for real-time ASR applications", Proc. of Int. Conf. Acoust., Speech, Signal Process., vol. 1, pp. 432-435, 2003.

Johnson, N.L., Kotz, S., & Balakrishnan, N., *Continuous Univariate Distributions*, John Willey and Sons. 1994.

Liu, H., Tang, Y. Q., & Zhang, H. H., "A new chi-square approximation to the distribution of non-negative definite quadratic forms in non-central normal variables", *Comput-ational Statistics and Data Analysis*, 53, 853 – 856, 2009.

Morris, A., Wu, D., & Koreman, J., "MLP trained to classify a small number of speakers generates discriminative features for improved speaker recognition", *Proceedings of IEEE Int. Carnahan Conference on Security Technology*, 11-14, 325 – 328, 2005.

Rabiner L., "A tutorial on hidden Markov models and selected applications in speech recognition", *Proceedings of the IEEE*, 77, 2, February, 1989.

Reynolds, D. A., Zissman, M. A., Quatieri, T. F., Leary, G. C. O., & Carlson, B. A., "The effect of telephone transmission degradations on speaker recognition performance", *Proceed-ings of International Conference on Acoustics, Speech and Signal Processing*, 329 – 332, 1995.

Reynolds, D. A., "Speaker identification and verification using Gaussian mixture speaker models", *Speech Commu-nication*, 17, 91 – 108, 1995.

Reynolds, D. A., "Large population speaker identification using clean and telephone speech", *IEEE Signal Processing Letters*, 2, 3, 46 – 48, 1995.

Reynolds, D., Quatieri, T. F., & Dunn, R. B., "Speaker Veri-fication Using Adapted Gaussian Mixture Models", *Digital Signal Processing*, 10, 1-3, 19 – 41, January, 2000.

S. M. Metev and V. P. Veiko, *Laser Assisted Micro-technology*, 2nd ed., R. M. Osgood, Jr., Ed. Berlin, Germany: Springer-Verlag, 2005.

Tierney, L., & Kadane. J., "Accurate approximations for posterior moments and marginal densities", *Journal of the American Statistical Association, 81(393)*, 1986.

Wu, D., Morris, A., & Koreman, J., "MLP internal repre-sentation as discriminative features for improved speaker recognition", *Nonlinear Analyses and Algorithms for Speech Processing Part II (series: Lecture Notes in Computer Science)*, 72 – 78, 2005.

Wu, D., Morris, A., & Koreman, J., "Discriminative features by mlp preprocessing for robust speaker recognition in noise", *Proc. of Electronic Speech Signal Processing*, 181 – 188, 2005.

Wu, D., Morris, A., & Koreman, J., "MLP internal representation as discriminative features for improved speaker recognition", *Proceedings of Nonlinear Speech Processing*, 25 – 32, 2005.

Wu, D., *Discriminative Preprocessing of Speech: Towards Improving Biometric Authentication*, Ph.D. thesis, Saarland University, Saarbruecken, Germany, July, 2006.

Wu, D., Li, B. J., & Jiang, H., "Normalisation and Transformation Techniques for Robust Speaker Recognition", *Speech Recognition: Techniques, Technology and Appli-cations*, V. Kordic (ed.), Springer Press, 1 – 21, 2008.

Wu, D., "Parameter Estimation for $\alpha$ -GMM based on Maximum Likelihood Criterion", *Neural Computation*, 21, 1 – 20, 2009.

Wu, D., Li, J., & Wu, H. Q., "$\alpha$ -Gaussian mixture models for speaker recognition", *Pattern Recognition Letters*, 30, 6, 589 – 594, 2009.

Young, S. et al., *HTKbook (V3.2)*, Cambridge University Engineering Dept., 2002.

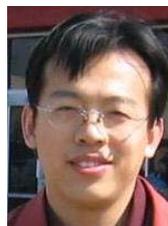 **Dalei Wu**- received B.S. degree in computer science from Harbin Engineering University in 1996, M.Sc. degree in Computer Science from Tsinghua University, China in 2002 and Ph.D. degree in Computational Linguistics from Saarland University, Saarbruecken, Saar-land, Germany in 2006.

During 2008-2010, he worked as a post-doctoral researcher in the department of computer science and engineering, York University. Since 2011, he has been working as a postdoctoral researcher in the department of electrical and com-puter engineering, Concordia University. His research interest focuses on automatic speech recognition, automatic speaker recognition and machine learning algorithms and speech enhance-ment.